\documentclass{article}

\usepackage{arxiv}

\usepackage[utf8]{inputenc} 
\usepackage[T1]{fontenc}    
\usepackage{hyperref}       
\usepackage{url}            
\usepackage{booktabs}       
\usepackage{amsfonts}       
\usepackage{nicefrac}       
\usepackage{microtype}      
\usepackage{cleveref}       
\usepackage{lipsum}         
\usepackage{graphicx}
\usepackage{natbib}
\usepackage{doi}
\usepackage{subcaption}

\title{The Latency Wall: Benchmarking Off-the-Shelf Emotion Recognition for Real-Time Virtual Avatars}

\date{}

\newif\ifuniqueAffiliation
\uniqueAffiliationtrue

\ifuniqueAffiliation 
\author{ \href{https://orcid.org/0009-0008-7427-6380}{\includegraphics[scale=0.06]{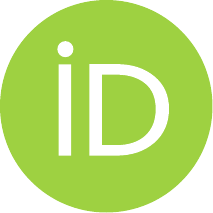}\hspace{1mm}Yarin Benyamin} \\
	Ben-Gurion University of the Negev\\
	\texttt{bnyamin@post.bgu.ac.il} \\
}

\renewcommand{\undertitle}{Technical Report}

\begin{document}
\maketitle

\begin{abstract}

In the realm of Virtual Reality (VR) and Human-Computer Interaction (HCI), real-time emotion recognition shows promise for supporting individuals with Autism Spectrum Disorder (ASD) in improving social skills. This task requires a strict latency-accuracy trade-off, with motion-to-photon (MTP) latency kept below 140 ms to maintain contingency. However, most off-the-shelf Deep Learning models prioritize accuracy over the strict timing constraints of commodity hardware. As a first step toward accessible VR therapy, we benchmark State-of-the-Art (SOTA) models for Zero-Shot Facial Expression Recognition (FER) on virtual characters using the UIBVFED dataset. We evaluate Medium and Nano variants of YOLO (v8, v11, and v12) for face detection, alongside general-purpose Vision Transformers including CLIP, SigLIP, and ViT-FER.Our results on CPU-only inference demonstrate that while face detection on stylized avatars is robust (100\% accuracy), a "Latency Wall" exists in the classification stage. The YOLOv11n architecture offers the optimal balance for detection ($\approx54$ ms). However, general-purpose Transformers like CLIP and SigLIP fail to achieve viable accuracy ($<23\%$) or speed ($>150 ms$) for real-time loops. This study highlights the necessity for lightweight, domain-specific architectures to enable accessible, real-time AI in therapeutic settings.

\end{abstract}


\section{Introduction}
The appearance of virtual avatars has revolutionized the landscape of human-computer interaction, with diverse domains ranging from gaming and entertainment to education and telecommunication. Embodying a wide array of forms, from humanoid figures to stylized characters and anthropomorphic entities, virtual avatars play a pivotal role in mediating social interactions, conveying emotions, and facilitating communication in virtual spaces.

Emotions play a fundamental role in human communication \citep{adolphs2002recognizing}. Individuals rely on subtle facial expressions, body language, and vocal cues to interpret the emotional states of others. 
By simulating virtual environments and leveraging machine learning algorithms, individuals can immerse themselves in scenarios where they learn to discern between a multitude of emotions.

Our primary focus is tackling the difficulties individuals with autism face when it comes to accurately understanding facial expressions\citep{celani1999understanding, lozier2014impairments}, a crucial skill for navigating social interactions. While one approach involves using wearable glasses to assist in recognizing these expressions\citep{haber2020making}\citep{elsherbini2023towards}, our research takes a different route. We aim to eliminate the necessity for wearable devices by providing a supportive and accessible virtual environment where individuals can train to improve their facial expression interpretation skills.

Regarding system latency, research indicates that while neurotypical individuals have a temporal binding window of approximately 300 ms, individuals with autism display an extended window of approximately 600 ms\citep{foss2010extended}. In Virtual Reality, neurotypical performance degrades significantly when latency exceeds roughly 70 ms\citep{caserman2019effects}. Given the twofold increase in the temporal binding window observed in autism, we posit that our system can target a proportionally relaxed latency budget of approximately 140 ms.
This threshold is not a clinically validated requirement but a conservative engineering heuristic informed by prior work, not a medical or therapeutic guarantee.

As a first step toward this goal, we benchmark state-of-the-art (SOTA) models for zero-shot emotion recognition on virtual characters using the UIBVFED dataset. In this evaluation, we do not purely seek the highest theoretical accuracy; rather, we analyze the critical trade-off between recognition precision and computational latency. To maintain the perceptual binding essential for our target demographic, the model's inference time must strictly adhere to our derived ~140 ms budget. Therefore, we investigate which architectures offer the optimal balance, ensuring sufficiently accurate expression analysis while remaining lightweight enough to prevent the performance degradation observed in high-latency VR environments.
This work does not propose a new model or therapy. It benchmarks existing off-the-shelf models against a latency budget, showing a key barrier to real-time emotion recognition in accessible VR.

\section{Methodology}

\subsection{Dataset \& Task Definition}

\begin{figure}[ht]
    \centering
    \begin{subfigure}[b]{0.23\textwidth}
        \centering
        \includegraphics[width=\textwidth]{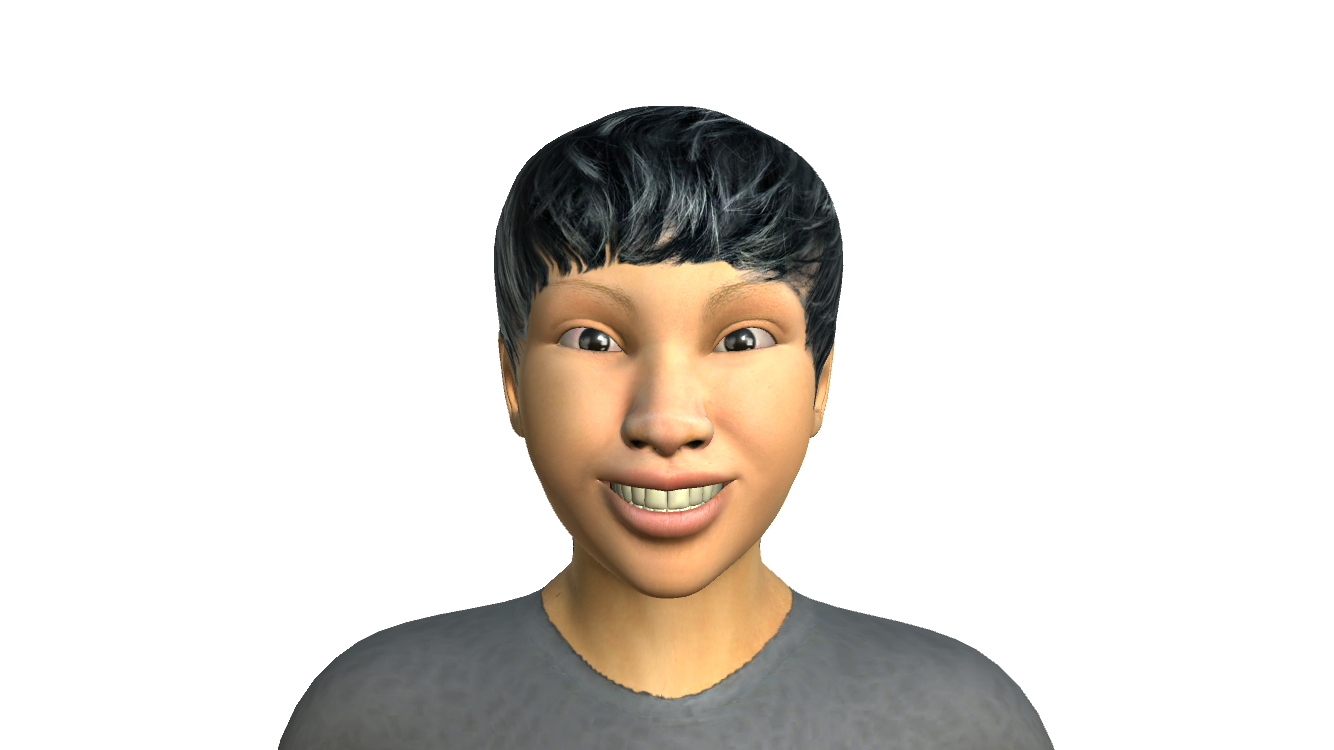}
    \end{subfigure}
    \hfill 
    \begin{subfigure}[b]{0.23\textwidth}
        \centering
        \includegraphics[width=\textwidth]{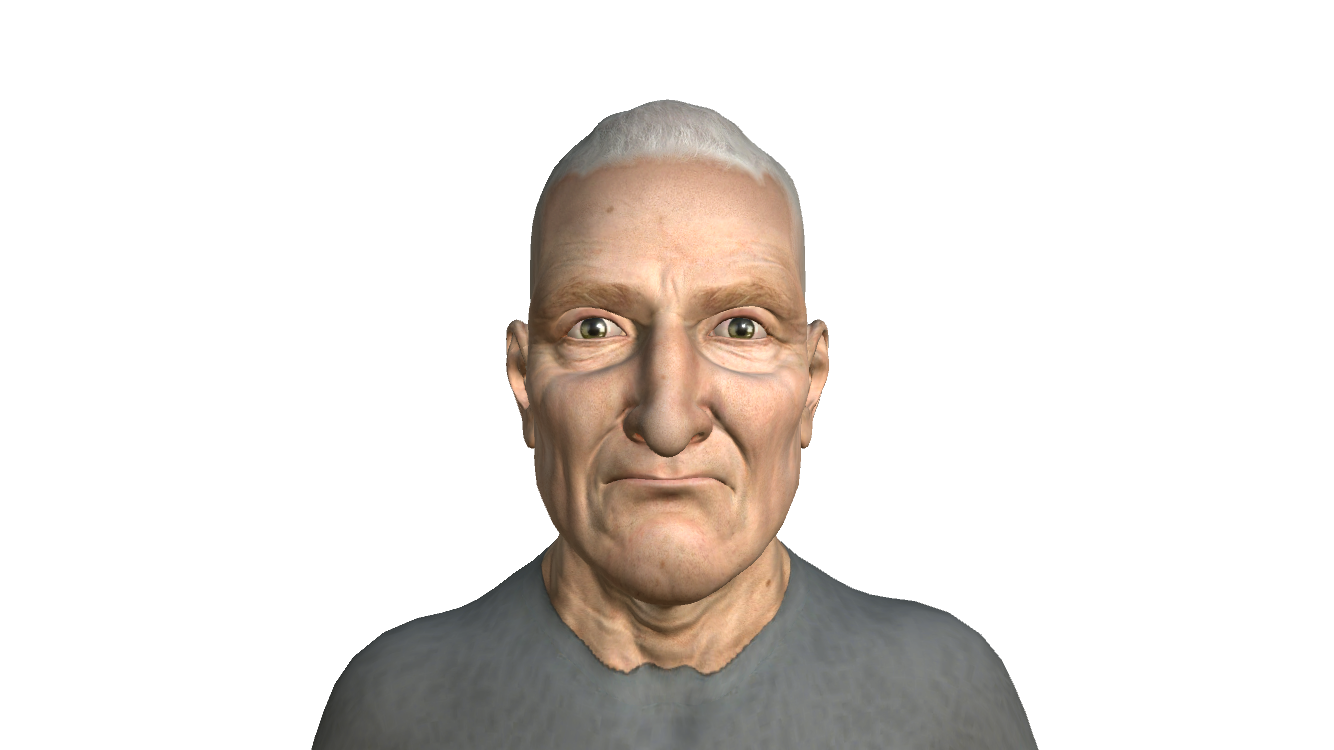}
    \end{subfigure}
    \hfill
    \begin{subfigure}[b]{0.23\textwidth}
        \centering
        \includegraphics[width=\textwidth]{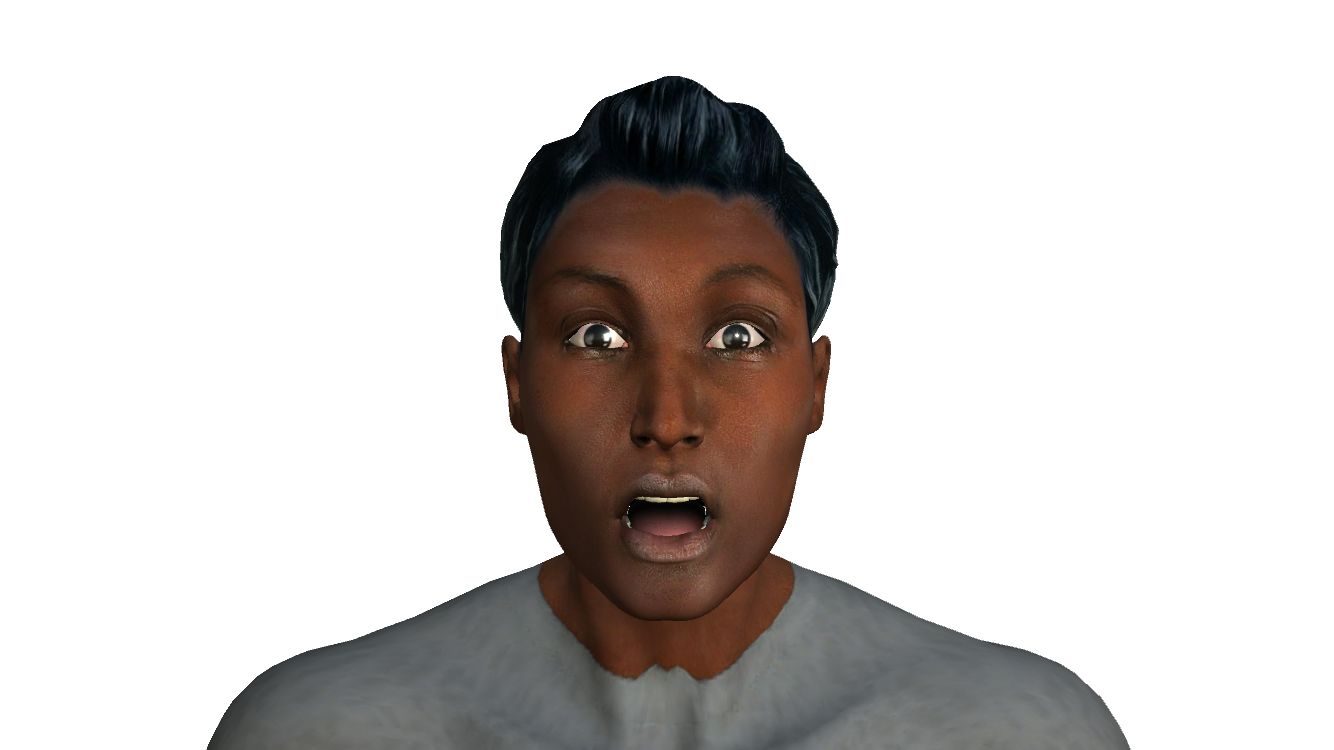}
    \end{subfigure}
    \hfill
    \begin{subfigure}[b]{0.23\textwidth}
        \centering
        \includegraphics[width=\textwidth]{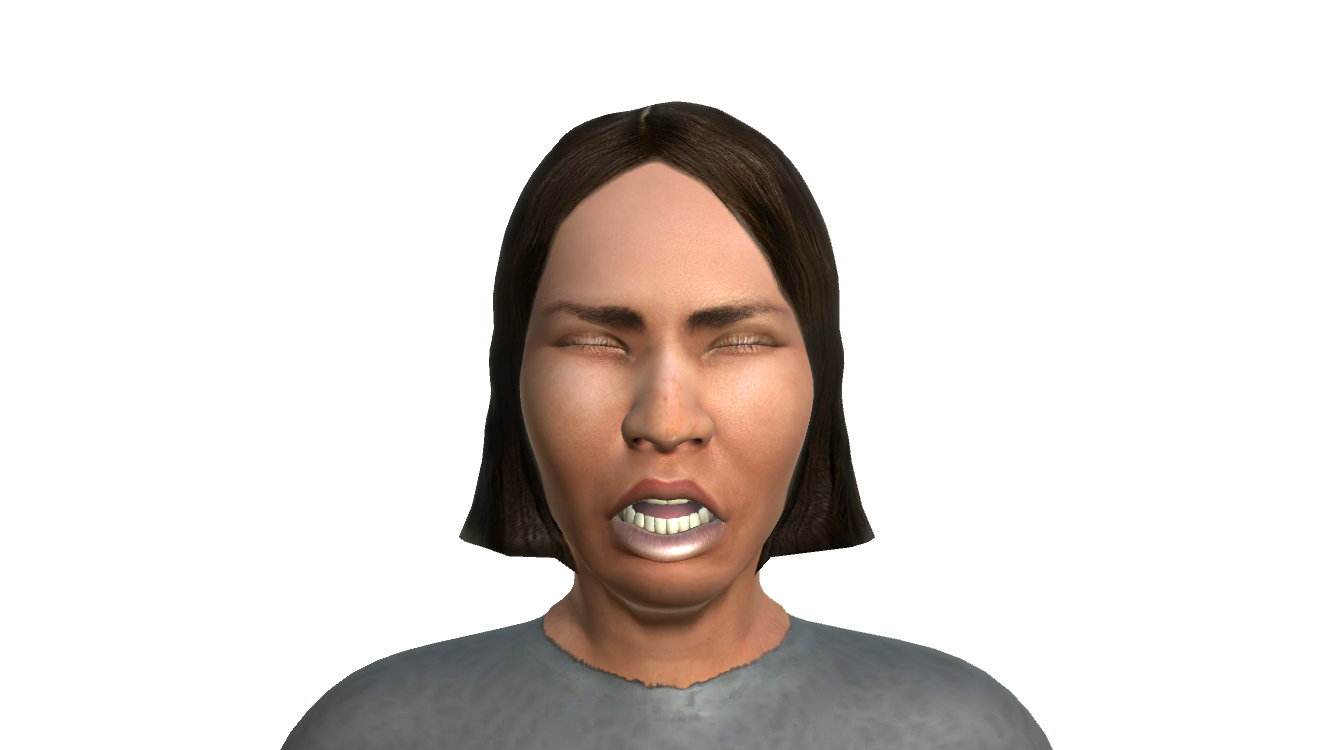}
    \end{subfigure}
    
    \caption{Images from the UIBVFED dataset}
    \label{fig:uibvfed_examples}
\end{figure}

The primary objective of this study is to evaluate the feasibility of real-time emotion recognition for virtual avatars. We define a standard set of seven categorical emotions: \texttt{["ANGER", "DISGUST", "FEAR", "JOY", "NEUTRAL", "SADNESS", "SURPRISE"]}.

To benchmark performance in the domain of non-photorealistic avatars, we utilize the \textbf{UIBVFED} dataset \citep{oliver2020uibvfed}, a collection of virtual facial expressions (see Figure \ref{fig:uibvfed_examples}). This dataset serves as a proxy for the stylized characters commonly used in VR therapy, allowing us to test domain generalization without training on specific character rigs.

\subsection{Pipeline Architecture}
We adopt a two-stage modular pipeline consisting of face detection followed by emotion classification. This design choice reflects the need to handle multiple faces within a single image while preserving clear identity-to-emotion associations. Rather than relying on end-to-end image-level classification, the proposed approach enables localized emotion inference at the individual face level. To facilitate rapid prototyping and establish strong baselines, we benchmark off-the-shelf models in a zero-shot or pre-trained setting, allowing us to assess their immediate applicability without task-specific fine-tuning.

\subsubsection{Stage 1: Face Detection}
We evaluate the \textbf{YOLO (You Only Look Once)} family of object detectors, specifically versions v8, v11, and v12 \citep{jocher2023yolo, tian2025yolov12}. To investigate the trade-off between accuracy and inference latency, we compare two model scales for each version:
\begin{itemize}
    \item \textbf{Medium (m):} Represents the standard balance of speed and accuracy.
    \item \textbf{Nano (n):} Highly optimized, lightweight architectures designed for mobile and edge deployment.
\end{itemize}

\subsubsection{Stage 2: Emotion Classification}
We assess three distinct architectures to determine the most effective approach for classifying stylized emotions:

\paragraph{Contrastive Language-Image Pre-training (CLIP \& SigLIP):} 
We utilize \textbf{CLIP} \citep{radford2021learning} and \textbf{SigLIP2} \citep{tschannen2025siglip} as Zero-Shot classifiers (in two model scales). These models are trained to align image and text representations in a shared latent space. 
For classification, we construct text prompts corresponding to our emotion labels (e.g., "ANGER"). We compute the cosine similarity between the image embedding and each text embedding, assigning the label with the highest similarity score:
    $\hat{y} = {argmax}_{c \in C} \left( \cos(\mathbf{I}, \mathbf{T}_c) \right)$
where $\mathbf{I}$ is the image embedding and $\mathbf{T}_c$ is the text embedding for class $c$.

\paragraph{ViT-FER (Domain Transfer):}
We also evaluate \textbf{ViT-FER} \citep{trpakov2023vit}, a Vision Transformer explicitly fine-tuned on the \textbf{FER-2013} dataset \citep{goodfellow2013challenges}. Unlike CLIP, this model is a dedicated classifier trained on human facial expressions. Including this model allows us to test the "Domain Gap", specifically, whether a model learned on human biometrics can generalize to the exaggerated features of virtual characters without fine-tuning.

\subsection{Experimental Setup}
To evaluate the accessibility of these tools for real-world therapeutic settings, where high-end GPUs are often unavailable, we conducted all benchmarks on a standard commercial workstation using \textbf{CPU-only inference}. The evaluation system was a Pop!\_OS 22.04 LTS (Linux) machine with a 12th Gen Intel® Core™ i7-1265U CPU (10 cores, 12 threads) and 32.0 GiB of RAM, reflecting the hardware constraints faced by clinicians or schools deploying portable VR interventions. Latency measurements represent the average end-to-end inference time over five runs.




\section{Results}

In this section, we present the benchmarking results divided by dataset domain. We first evaluate the primary target domain (Stylized Virtual Avatars / UIBVFED) to assess system feasibility, followed by the human baseline (FER-2013) to evaluate domain generalization and architecture stability.

\subsection{UIBVFED Benchmark (Virtual Avatars)}
This benchmark represents the ``Output'' stage of the pipeline, verifying if the system can accurately perceive the stylized expressions of the virtual avatar itself.

\subsubsection{Face Detection on UIBVFED}
As shown in Table \ref{tab:uibvfed_detection}, detection robustness on high-resolution virtual avatars was absolute, with all architectures achieving \textbf{100\% accuracy}. This shifts the critical evaluation metric entirely to latency. 

A notable finding is the performance of the legacy \textbf{MTCNN}. While accurate, it suffered from a ``Resolution Trap,'' slowing to 78.78 ms on high-res inputs, making it significantly slower than modern Nano architectures. \textbf{YOLOv11n} proved to be the optimal choice for this domain, offering the lowest latency ($\approx$54 ms).

\begin{table}[h]
\centering
\caption{Face Detection Results on UIBVFED (Virtual Avatars)}
\label{tab:uibvfed_detection}
\begin{tabular}{@{}llcc@{}}
\toprule
\textbf{Model} & \textbf{Size} & \textbf{Accuracy} & \textbf{Latency (ms)} \\ \midrule
YOLOv8m & Medium & 100.00\% & 448.16 \\
YOLOv11m & Medium & 100.00\% & 229.18 \\
YOLOv12m & Medium & 100.00\% & 242.90 \\ \midrule
YOLOv8n & Nano & 100.00\% & 56.68 \\
\textbf{YOLOv11n} & \textbf{Nano} & \textbf{100.00\%} & \textbf{53.96} \\
YOLOv12n & Nano & 100.00\% & 62.72 \\ \midrule
MTCNN & Baseline & 100.00\% & 78.78 \\ \bottomrule
\end{tabular}
\end{table}

\subsubsection{Emotion Classification on UIBVFED}
Despite perfect detection, emotion classification remains the primary bottleneck (Table \ref{tab:uibvfed_classification}). Zero-shot models failed to transfer effectively to the virtual domain. \textbf{CLIP-Large} provided the highest zero-shot accuracy (22.88\%) but at a prohibitive latency cost ($\approx$1.7s). \textbf{ViT-FER}, trained on human data, achieved the best trade-off (27.42\%), yet its low accuracy confirms a substantial domain gap that off-the-shelf models cannot currently bridge.

Crucially, with a 7-class problem space, a random guess yields $\approx14.2\%$ accuracy. Consequently, models like SigLIP (4.24\%) performed worse than random chance, and even the best SOTA models barely exceeded this baseline, indicating a fundamental failure in domain transfer, not just latency.

\begin{table}[h]
\centering
\caption{Emotion Classification Results on UIBVFED (Virtual Avatars)}
\label{tab:uibvfed_classification}
\begin{tabular}{@{}llcc@{}}
\toprule
\textbf{Model} & \textbf{Variant} & \textbf{Accuracy} & \textbf{Latency (ms)} \\ \midrule
CLIP & ViT-B/16 & 15.91\% & 153.60 \\
CLIP & ViT-L/14 & 22.88\% & 1730.94 \\ \midrule
SigLIP2 & Base & 9.09\% & 180.14 \\
SigLIP2 & So400m & 4.24\% & 3097.04 \\ \midrule
\textbf{ViT-FER} & \textbf{Base} & \textbf{27.42\%} & \textbf{193.58} \\ \bottomrule
\end{tabular}
\end{table}

\subsection{FER-2013 Benchmark (Real Human Faces)}
This benchmark represents the ``Input'' stage of the pipeline (perceiving the user) and serves as a control group to evaluate architectural stability.

\subsubsection{Face Detection on FER-2013}
Table \ref{tab:fer_detection} reveals a critical instability in newer YOLO architectures. While \textbf{YOLOv8n} maintained robust performance on human faces (80.79\%), the newer \textbf{YOLOv11n} and \textbf{YOLOv12n} models collapsed to 21.59\% and 35.18\%, respectively. Conversely, \textbf{MTCNN} demonstrated its strength on low-resolution human data, achieving the fastest global latency (8.3 ms).

This suggests that while v11 is superior for stylized virtual characters, \textbf{YOLOv8n} is the only ``Hybrid'' candidate capable of handling both human inputs and avatar outputs reliably. 
However, with an inference time of $\approx63$ms, YOLOv8n consumes nearly 50\% of the total 140ms budget on detection alone, leaving insufficient time for complex emotion classification.

\begin{table}[h]
\centering
\caption{Face Detection Results on FER-2013 (Human)}
\label{tab:fer_detection}
\begin{tabular}{@{}llcc@{}}
\toprule
\textbf{Model} & \textbf{Size} & \textbf{Accuracy} & \textbf{Latency (ms)} \\ \midrule
\textbf{MTCNN} & \textbf{Baseline} & \textbf{83.64\%} & \textbf{8.30} \\ \midrule
YOLOv8n & Nano & 80.79\% & 63.40 \\
YOLOv11n & Nano & 21.59\% & 67.10 \\
YOLOv12n & Nano & 35.18\% & 83.70 \\ \bottomrule
\end{tabular}
\end{table}

\subsubsection{Emotion Classification on FER-2013}
As expected, models pre-trained on human data performed significantly better in this domain (Table \ref{tab:fer_classification}). \textbf{ViT-FER} achieved 63.64\% accuracy, confirming that its poor performance on UIBVFED was due to domain shift rather than model incapacity. Similarly, CLIP-Base jumped from $\approx$16\% (Virtual) to $\approx$40\% (Human), illustrating the bias of large-scale pre-training datasets toward photorealistic human features.

\begin{table}[h]
\centering
\caption{Emotion Classification Results on FER-2013 (Human)}
\label{tab:fer_classification}
\begin{tabular}{@{}llcc@{}}
\toprule
\textbf{Model} & \textbf{Variant} & \textbf{Accuracy} & \textbf{Latency (ms)} \\ \midrule
CLIP & ViT-B/16 & 39.97\% & 151.60 \\
CLIP & ViT-L/14 & 38.51\% & 1673.70 \\ \midrule
SigLIP2 & Base & 17.11\% & 177.80 \\ \midrule
\textbf{ViT-FER} & \textbf{Base} & \textbf{63.64\%} & \textbf{189.80} \\ \bottomrule
\end{tabular}
\end{table}

\section{Discussion: The ``Latency Wall''}

\subsection{Total Pipeline Latency}
To achieve a ``Sense of Agency,'' the total loop must occur in under 140ms.
\begin{equation}
    T_{total} = T_{detect} + T_{classify} + T_{render}
\end{equation}
Assuming a minimal budget of 10ms for capture/rendering, our AI processing must occur within \textbf{130ms}.

Using our absolute fastest configuration (YOLOv11n + ViT-FER):
\begin{equation}
    T_{AI} = 53.96ms + 193.58ms = 247.54ms
\end{equation}
This result is \textbf{1.9x} the allowable budget.

This confirms our hypothesis: standard Vision Transformers, even in their "Base" configurations, are fundamentally too heavy for CPU-based real-time interaction loops required for autistic contingency. 
These benchmarks highlight the barrier to entry for accessible therapy. All results were obtained on a standard i7 Laptop CPU, replicating the hardware constraints of a typical clinic or school. The inability of any SOTA pipeline to meet the 140ms target on this hardware suggests that current AI-driven interventions are implicitly gated behind high-end GPU requirements, limiting their accessibility to well-funded institutions.

\subsection{Design Implications for VR and HCI Systems}
Our results suggest that real-time, contingency-sensitive avatars should not rely on large, general-purpose models, because they introduce unacceptable latency on commodity VR hardware. 
When response times exceed the $\approx$140 ms threshold, users may experience delayed social cues, which can weaken the sense of presence and even create negative reinforcement by associating interactions with lag. 
Therefore, designers should prioritize specialized, lightweight architectures and optimized pipelines, using techniques such as knowledge distillation and quantization, to achieve fast and reliable performance. 
This shift is necessary to balance high-quality perception with practical deployment, and it highlights the need for hardware-aware solutions rather than off-the-shelf transfer learning for accessible VR applications.

\section{Conclusion}
In this work, we evaluated the feasibility of off-the-shelf computer vision models for real-time, contingency-sensitive virtual avatars, revealing a distinct bifurcation in performance between detection and classification tasks. While face detection on stylized avatars proved robust, with all architectures achieving 100\% accuracy, we identified a critical trade-off in architectural specialization: \textbf{YOLOv11n} emerged as the optimal "Specialist" for virtual domains due to its superior speed, whereas \textbf{YOLOv8n} proved to be the more robust "Generalist" for mixed-reality applications involving human inputs. However, the system remains fundamentally constrained by a "Latency Wall" in emotion classification, where Zero-Shot and Transfer Learning models incur latencies ($\approx$250ms) that violate the sub-140ms motion-to-photon budget required for therapeutic agency. Consequently, we conclude that standard Transformer-based pipelines are ill-suited for commodity hardware, and future work must prioritize distilling these heavy architectures into lightweight, domain-specific CNNs to bridge the gap between high-fidelity perception and real-time social contingency.

\section*{Acknowledgments}
The author extends their gratitude to B.Sc. Yuval Lee Englander for the original concept of the therapeutic environment. While the environment itself was conceptualized during the course ``Virtual and Augmented Reality (VR \& AR) for Research, Treatment and Rehabilitation,'' led by Dr. Shachar Maidenbaum, the specific investigation into latency constraints presented in this report represents a technical extension of that foundation.

\bibliographystyle{unsrtnat}
\bibliography{references}  






\end{document}